%% file: main.tex
\documentclass[letterpaper, 10 pt, conference]{ieeeconf}  

\IEEEoverridecommandlockouts                              
\usepackage[utf8]{inputenc}
\usepackage[T1]{fontenc}

\usepackage{times}
\usepackage{epsfig}
\usepackage{graphicx}
\usepackage{amsmath, bm}
\usepackage{amssymb}
\usepackage{textcomp}
\usepackage{xcolor}
\usepackage{hyperref}
\usepackage[absolute]{textpos}

\title{\LARGE \bf
Predicting Signed Distance Functions for Visual Instance Segmentation 
}

\author{ \parbox{6 in}{\centering Emil Brissman$^{1,2}$, Joakim Johnander$^{1,3}$, Michael Felsberg$^{1}$\\
         \vspace{0.1 in}
         $^{1}$Computer Vision Laboratory, Dept. of Electrical Engineering, Linköping University \\
         $^{2}$Saab, Sweden\\
         $^{3}$Zenseact, Sweden\\
         {\tt\small \{emil.brissman, joakim.johnander, michael.felsberg\}@liu.se}}
}

\begin{document}

\maketitle
\thispagestyle{empty}
\pagestyle{empty}

\begin{abstract}

Visual instance segmentation is a challenging problem and becomes even more difficult if objects of interest varies unconstrained in shape. Some objects are well described by a rectangle, however, this is hardly always the case. Consider for instance long, slender objects such as ropes. Anchor-based approaches classify predefined bounding boxes as either negative or positive and thus provide a limited set of shapes that can be handled. Defining anchor-boxes that fit well to all possible shapes leads to an infeasible number of prior boxes. We explore a different approach and propose to train a neural network to compute distance maps along different directions. The network is trained at each pixel to predict the distance to the closest object contour in a given direction. By pooling the distance maps we obtain an approximation to the signed distance function (SDF). The SDF may then be thresholded in order to obtain a foreground-background segmentation. We compare this segmentation to foreground segmentations obtained from the state-of-the-art instance segmentation method YOLACT. On the COCO dataset, our segmentation yields a higher performance in terms of foreground intersection over union (IoU). However, while the distance maps contain information on the individual instances, it is not straightforward to map them to the full instance segmentation. We still believe that this idea is a promising research direction for instance segmentation, as it better captures the different shapes found in the real world.

\end{abstract}

\input{introduction}


\input{relatedwork}

\input{method}

\input{results}

\section{ACKNOWLEDGEMENT}
This work was partially supported by the Wallenberg AI, Autonomous Systems, and Software Program (WASP) funded by the Knut and Alice Wallenberg Foundation.

\bibliographystyle{ieee}
\bibliography{egbib}

\end{document}

%% file: introduction.tex
\section{INTRODUCTION}

Instance segmentation is a core computer vision problem. Given an image, the aim is to detect and segment all objects in an image. A class label is associated to each pixel in the image, similar to semantic segmentation, except that multiple objects of the same class shall be treated as separate entities. This problem has been extensively studied in recent years but remains challenging. The dominant approaches rely on \emph{anchorboxes}~\cite{Bolya_2019_ICCV,He_2017_ICCV}. Anchorboxes is a predefined set of axis-aligned bounding boxes with one or a few different aspect ratios. The idea is that each object shape and object location in the scene should be approximately described by one or more anchorboxes. These approaches then train a neural network to classify each anchorbox as an object or as background. Usually, most objects are fairly well described by one or more of these anchorboxes. However, several common objects such as ropes or road lanes, are not well described by this approach.

In this work, we explore a new direction with the aim to tackle instance segmentation. The idea is to at each pixel predict the distance to the closest object contour, in a given direction. We use Signed Distance Functions (SDFs), giving us a positive distance if we are within an object and a negative distance if we are outside an object. By thresholding, we immediately know whether we are within an object or not. Furthermore, the object contours are found at the zero-crossings of the SDF. In practice, the SDF will deviate slightly; negative if the contour is too far inside and positive if it is too far outside. This is like a spring model and the global equilibrium determines the location of the contour. By representing detected objects with the SDF, avoiding the predefined set of anchorboxes, we are able to capture objects of arbitrary shapes.


Our contributions are
\begin{itemize}
    \item We propose to represent segmentations with signed distance functions
    \item We show that standard encoder-decoder neural networks are able to learn to predict these functions
    \item We quantitatively demonstrate that using the signed distance function output, we are able to produce a foreground-background image segmentations of quality comparable to what is obtained with state-of-the-art instance segmentation algorithms
\end{itemize}
We are able to efficiently get both the foreground and the contours of objects. An efficient way to separate the different instances from the signed distance function is subject to future work.

\begin{figure}[t]
\vspace{0.09 in}
\begin{center}
    \includegraphics[width=0.5\linewidth]{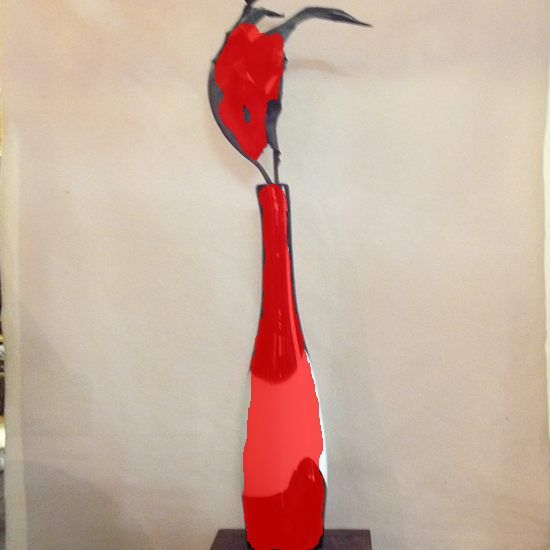}\includegraphics[width=0.5\linewidth]{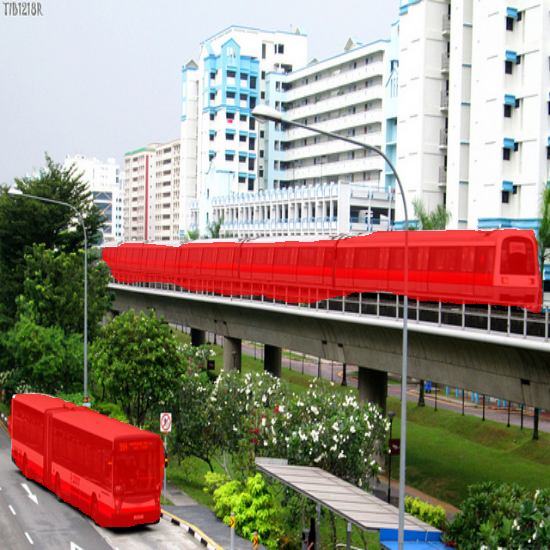}
\end{center}
   \caption{We show two example segmentations predicted by our method. In both cases, we have long, slender objects that are not well described by anchor-boxes. Our network instead predicts, for each pixel, the signed distance to the closest contour along a set of directions. By taking the sign of these functions, we directly obtain a segmentation of the input image.}
\label{fig:3d}
\end{figure}
\pubidadjcol 

%% file: relatedwork.tex
\begin{figure*}[t!]
\begin{center}
    \includegraphics[width=0.99\linewidth]{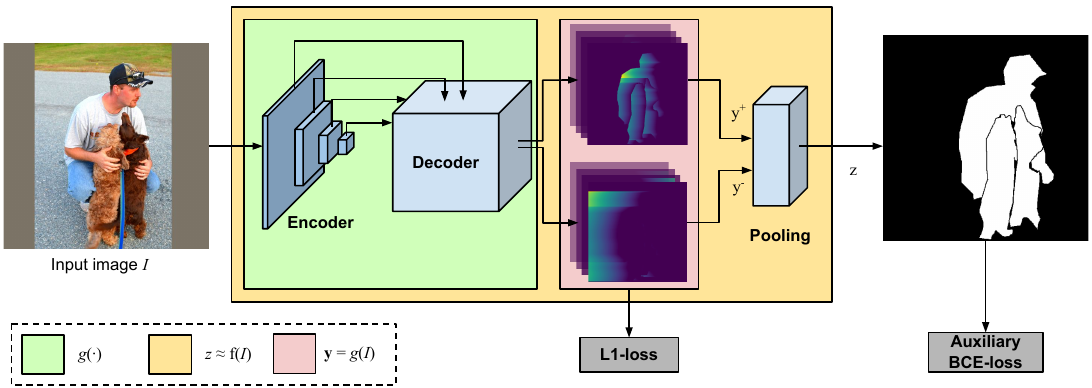}
\end{center}
   \caption{Overview of our approach. The encoder-decoder network outputs distance maps along the directions of $\theta$. The output $\mathbf{y}$ is divided into a positive part $y^+\in\mathbb{R}^{4\times H\times W}$ that defines the foreground and a corresponding negative part, that defines the background. We apply min-pooling to the parts separately and subtract the positive part with the negative. After subtraction, $z \approx f(\mathcal{I})$ and the foreground segmentation is directly obtained by taking the sign of $z > 0$.}
\label{fig:model}
\end{figure*}

\section{RELATED WORK}
\label{sec:related}
A lot of research work has been put down into increasing the accuracy of instance segmentation and a lot of work is still ongoing. In three sections we review related methods (Section~\ref{sec:related:ssis}) and highlight alternative approaches to the instance segmentation problem (Sections~\ref{sec:related:embedding} to \ref{sec:related:sdf}).

\subsection{Anchor-boxes}
\label{sec:related:ssis}
Instance segmentation approaches that rely on anchor-boxes still dominate, to a large extent, the baseline of state-of-the art methods. Mask R-CNN~\cite{He_2017_ICCV} is anchor-based and splits the task into two subtasks, first object detection then object segmentation. Subsequent computations restrict this approach for real-time speed. Single-stage instance segmentation, like YOLACT~\cite{Bolya_2019_ICCV}, improve this aspect but yield lower accuracy. This work directly predicts pixel position representations that are assembled into final results. For segmentation, Bolya et al.~\cite{Bolya_2019_ICCV} suggest to compute object foreground as a logistic regression problem where regression hyper parameters are inferred at each pixel position. In contrast to our approach we do not have any prior assumption on position or shape for the instance segmentation task. 

\subsection{Spatial embeddings}
\label{sec:related:embedding}
Subsequent methods attempts to perform instance segmentation without anchor-boxes and a popular branch is proposal-free methods. CenterNet~\cite{zhou2019objects} suggest to model each object as its centerpoint. Similarly, but for the instance segmentation task, CenterMask~\cite{wang2020centermask} propose to adopt local shape information from the representation of object centerpoints to compute the object segmentations. In Neven et al.~\cite{Neven_2019_CVPR}, segmentations are acquired by clustering pixel embeddings that assigns a pixel to an object. A major drawback is that objects could acquire the same centerpoint, which makes it hard to discriminate between the instances. A recent method~\cite{Carion_ECCV_2020} considers all the pixels as object queries from which self-attention retain those with coherent activations. The top remaining queries are subsequently assigned as background or foreground, and in the latter case, class and bounding box are predicted.



\subsection{Signed distance function}\label{sec:related:sdf}
TensorMask~\cite{Chen_2019_ICCV} created a foundation for exploring new directions for research within instance segmentation. Chen et al.~\cite{Chen_2019_ICCV} motivated to building a strong image understanding from structured 4D tensors to represent complex object shapes. This inspired our work to consider signed distance functions. Such functions are simplistic but able to model complex object shapes. In 3D-graphics the signed distance function show more frequent usage, like in previous work~\cite{Liu_2020_CVPR,Park_2019_CVPR}, but that only considers a single object.


%% file: method.tex
\begin{figure*}[ht]
\begin{center}
    \includegraphics[width=0.49\linewidth]{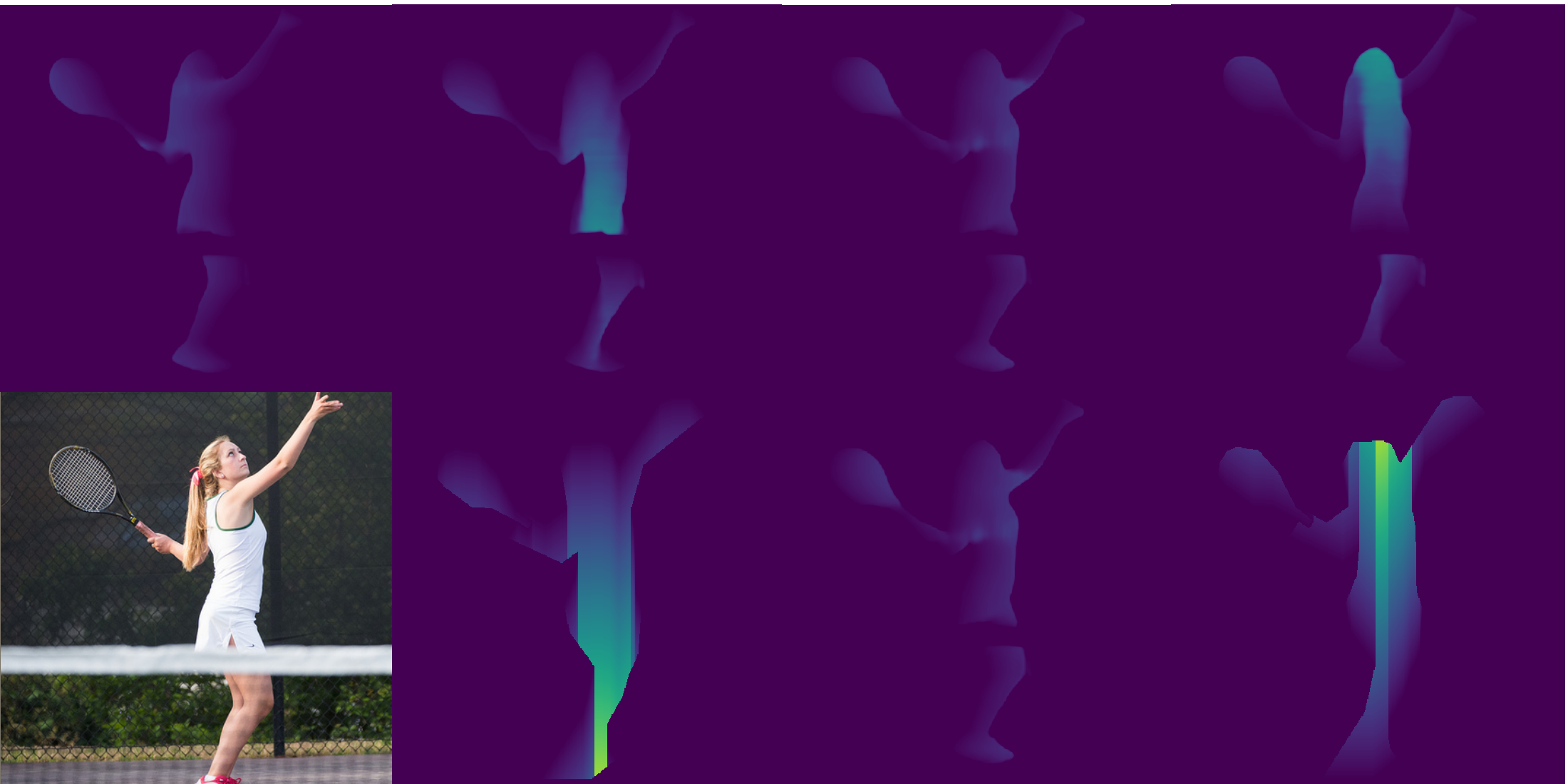}\hspace{0.08 in}\includegraphics[width=0.49\linewidth]{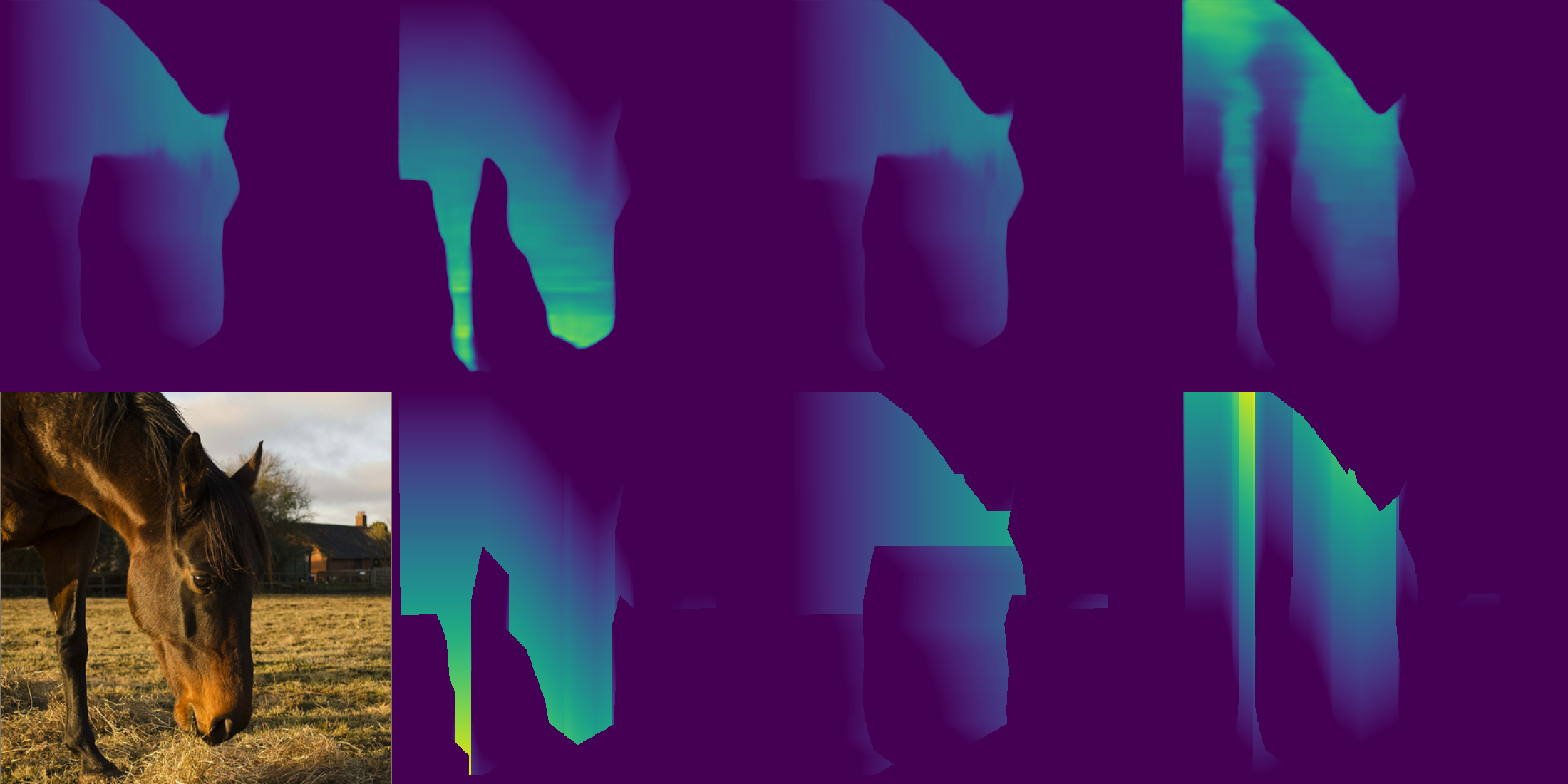}
\end{center}
   \caption{We show two examples by the positive part, $y^+\in\mathbb{R}^{4\times H\times W}$, along the directions $\{0, 90, 180, 270\}$. The leftmost image depicts a woman playing tennis and the rightmost image depicts a horse. The bottom row show the original image and the three last distance annotations, $\{90, 180, 270\}$. Further details about the generation of distance annotations can be found in Section~\ref{sec:impl}. Although the ground truth neglects the tennis net that occludes the woman who plays tennis, the distance predictions do not.}
\label{fig:dist}
\end{figure*}

\section{METHOD}

We aim to develop a neural network that learns to approximate a Signed Distance Function (SDF) and then use the SDF for visual instance segmentation. The trained network should at each pixel predict the distance to the closest object contour in a set of different directions. These distances provide the information necessary to segment object instances, and the image is easily segmented into foreground and background by thresholding the SDF. An overview of the approach is shown in Fig.~\ref{fig:model}.

\subsection{Distance maps}
A signed distance function $f:\mathbb{R}^2\rightarrow\mathbb{R}$ is a continuous function that, for a given pixel position $\mathbf{p}\in\mathbb{R}^2$ in an image $\mathcal{I}$, outputs the distance to the closest object contour. The function sign encodes whether $\mathbf{p}$ is on the object (positive) or outside of the object (negative). Furthermore, all underlying object contours are implicitly represented by the iso-curves $f\left( \cdot \right) = 0$. The SDF, $f$, is defined as
\begin{align}
    f(\mathbf{p}) = s\cdot \min_{d\in\mathcal{D}_\theta^\mathbf{p}} d\enspace, \label{eq:sdf}
\end{align}
where
\begin{align}
    \mathcal{D}_\theta^\mathbf{p} = \{\infty\}\cup \{d\in\mathbb{R}^+: \mathbf{p} + d\begin{pmatrix}\cos(\theta) \\ \sin(\theta)\end{pmatrix} \in \Omega\}\enspace. \label{eq:distset}
\end{align}
Here, $d$ represents distances to contours in the direction $\theta$ and $s\in\{-1, 1\}$ is the sign. If $\mathbf{p}$ is within an object we let $s$ be positive, and otherwise negative. The set $\Omega$ contains a subset of image pixel positions, each of which belongs to object contours. That is, $\mathcal{D}_\theta^\mathbf{p}$, in \eqref{eq:distset}, contains the distances to object contours that are intersected by the lines $\mathbf{p} + d\begin{pmatrix}\cos(\theta)&\sin(\theta)\end{pmatrix}^T$. The signed distance value is the minimum of this set, see \eqref{eq:sdf}. The set will only contain the value of infinity if there is no intersection for each of the directions. In other words, the network encodes the distance to the closest object contour in a discrete number of directions. Since the number of directions is discrete, we consider the SDF in \eqref{eq:sdf} as an approximation. See Fig.~\ref{fig:dist} for an example on the output maps.






\subsection{Our approach}
\label{sec:netarch}
Similar to semantic segmentation, we want to predict a vector that encodes distances to the closest object contours for each pixel. That is, given an input image, the output $\mathbf{y} = g(\mathcal{I})$ encodes the geometric structures established by the training dataset. For this purpose we use an encoder-decoder network $g:\mathbb{R}^{3\times H\times W}\rightarrow \mathbb{R}^{2\times4\times H\times W}$ that predicts distances in four directions of $\theta_k = 90k$ degrees, where $k\in\{0,\:\hdots,\:3\}$. The output $\mathbf{y}$ is divided into two parts. A negative part $y^-\in\mathbb{R}^{4\times H\times W}$ that defines the background, and a corresponding positive part, $y^+$, that defines the foreground. To compute an approximate SDF $z_{uv} \approx f(\mathbf{p})$, we use
\begin{align}
    z_{uv} = \min_{k}y_{kuv}^{+} - \min_{k}y_{kuv}^{-}\enspace,
    \label{eq:pool}
\end{align}
where $\mathbf{p} = \begin{pmatrix}u&v\end{pmatrix}^T$. In other words, we take the minimum value along the first dimension of both parts and subtract these for all pixels.

We apply a $L_1$-loss to the distance output of $g$, such that
\begin{align}
    \mathcal{L}_{\text{dist}} = \sum_{(u,v) \in \mathcal{I}}\:\:\sum_{\forall k} \vert \hat{y}_{kuv}^+ - y_{kuv}^+ \vert + \vert \hat{y}_{kuv}^- - y_{kuv}^- \vert \label{eq:loss}
\end{align}
is minimised. In~\eqref{eq:loss} we denote $y^+$ as the annotated foreground distance and $\hat{y}^+$ as the predicted foreground distance. Similarly for the background part of~\eqref{eq:loss}.

In $\mathcal{L}_{\text{dist}}$ we utilise the value zero for two purposes, 1) to minimise the distance between the predicted distance and the true distance to an object contour and 2) for invalid distances, that is, when $\mathcal{D}_\theta^\mathbf{p}$ contains the value of infinity. From a pixel position that in a given direction does not intersect any object contours is in fact invalid. To use zero in the latter case maintains a simple loss. For the $L_1$-loss this means that the expected value for each invalid location with a predicted positive distance should be suppressed to zero.

We initially observed that the predicted distances were noisy. We therefore regularise the network by adding a binary cross-entropy loss $\mathcal{L}_{\text{aux}}$ to $z$. This loss drives the network to predict correct foreground-background segmentations, consistent with the distance output $\mathbf{y}$. We sum the two sub-losses into the total loss $\mathcal{L}_{\text{tot}} = \mathcal{L}_{\text{dist}} + \mathcal{L}_{\text{aux}}$ and backpropagate.

\subsection{Implementations details}
\label{sec:impl}
We train the model, including the encoder, with batch size 8 on one GPU using the COCO 2017 benchmark~\cite{lin2014microsoft}. This dataset is divided into 118K training samples and 5K validation samples. The encoder used ImageNet~\cite{deng2009imagenet} pretrained weights and the decoder was randomly initialised sampling from a uniform distribution. We train with SGD for 54 epochs starting at an initial learning rate of $2*10^{-3}$ and divide with 10 reaching epochs 19, 40, 48 and 51. Weight decay and momentum was $10^{-4}$ and $0.9$ respectively. For data augmentation we used the same as in SSD~\cite{Liu2016SSDSS}, also used by YOLACT~\cite{Bolya_2019_ICCV}. Training takes 4 days on a single Nvidia V100 GPU.

\subsubsection{Encoder-decoder}
In our model we use a ResNet50~\cite{He2016DeepRL} feature extractor as the encoder and a DFN~\cite{lin2014microsoft} as the decoder. DFN~\cite{lin2014microsoft} predicts a high-resolution output by fusing the deep feature maps with successively shallower features. Our distance model is motivated by this aspect. Hence, we hypothesise that consistency are kept for distance predictions to the contours of multi-scale objects. In the pooling layer the predicted distance maps are heuristically merged using two min-pooling operations that are concatenated with the correct sign $s$, for background and foreground. Finally, this is summed along the first dimension, see ~\eqref{eq:pool}.


%

\subsubsection{Distance annotations}
\label{sec:anno}
COCO~\cite{lin2014microsoft} comprises 80 classes of objects. These classes were selected to well represent frequently used words to describe visually identifiable objects. The selection is based on the PASCAL VOC dataset~\cite{pascalvoc}, a list of the 1200 most common words to describe objects, where several children between 4 and 8 were asked to name every object encountered in indoor and outdoor environments~\cite{lin2014microsoft}. The result is a set of 80 distinct categories, most fairly high-level, e.g. car, human, or dog.
In each image, each object is annotated with a segmentation mask. Based on these masks, we compute the index map. In some cases, multiple object masks overlap, leading to some pixels being claimed by multiple objects. Our distance annotations rely on each pixel mapping the distance to the single closest object contour. We therefore, for those pixels, select the single object mask corresponding to the smallest mask of all the overlapping object masks. The rationale is that we then give all objects a chance to appear. This can of course turn out wrong in certain cases. As in the upper left image of Fig. ~\ref{fig:gb}, where people in the background, which are occluded by people at the front, will be prioritised incorrectly.

Distance maps, used in the loss which is described in Section~\ref{sec:netarch}, are computed for the directions $\{0, 90, 180, 270\}$. We first rotate the index map according to the directions. This enables row-wise computation of the distance to the closest object contour. Finally the distance maps are rotated back to their initial state. The procedure is vectorised to increase overall GPU utilisation, which means faster training.




%% file: results.tex
\section{RESULTS}
\label{sec:results}

We report a result on COCO instance segmentation ~\cite{lin2014microsoft}. Our model as, well as YOLACT~\cite{Bolya_2019_ICCV}, is trained on the 118K ~\texttt{train2017} images. We test out approach on the 5K \texttt{val2017} images and compare our foreground segmentation with the foreground segmentation obtained from YOLACT~\cite{Bolya_2019_ICCV}. We use the intersection-over-union averaged over all 5K samples in \texttt{val2017}.

Table~\ref{tab:iou} compares the IoU of the foreground segmentation by \emph{topk} detections and changing thresholds on $z$. Our approach yields a $4\%$ increase over the state-of-the-art method YOLACT~\cite{Bolya_2019_ICCV}. This result demonstrates that managing of complex shapes can be beneficial. We also observe a negative effect when increasing the number of \emph{topk} detections as well as for larger and larger thresholds on $z$. However, the latter is expected, since small objects should disappear as result of the distance modelling. The former is contradictory towards this anchor-based method, since it should be expected that performance would go up when enabling more objects to be detected. 

Using our method to predict foreground segmentation is faster compared to YOLACT~\cite{Bolya_2019_ICCV}. We use the same feature extractor as in Bolya et al.~\cite{Bolya_2019_ICCV}, but without a computationally expensive Feature Pyramid Network~\cite{Lin_2017_CVPR}.  However, this speed is not fair to compare, since more computations in our approach is necessary to achieve full instance segmentation.

In Fig.~\ref{fig:dist} we show two qualitative examples of our predicted distances. The rightmost show a simple example of a hoarse where prediction and ground truth are comparable. In the leftmost example a woman is playing tennis. The distance predictions are harder to compare to the ground truth since our model finds that the woman is occluded by a tennis net.

\begin{figure*}[ht]
\begin{center}
    \includegraphics[width=0.33\linewidth]{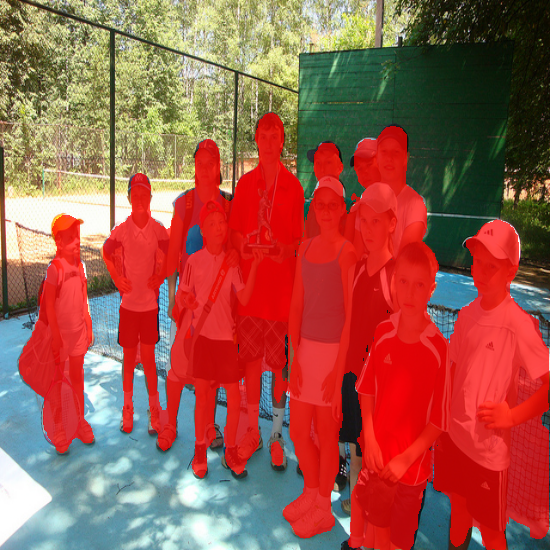}\includegraphics[width=0.33\linewidth]{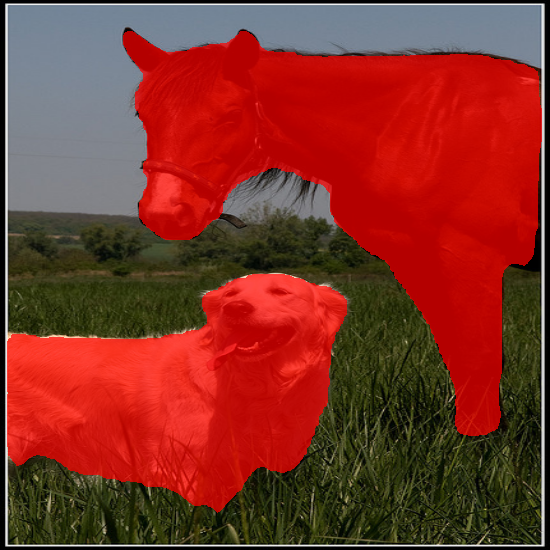}\includegraphics[width=0.33\linewidth]{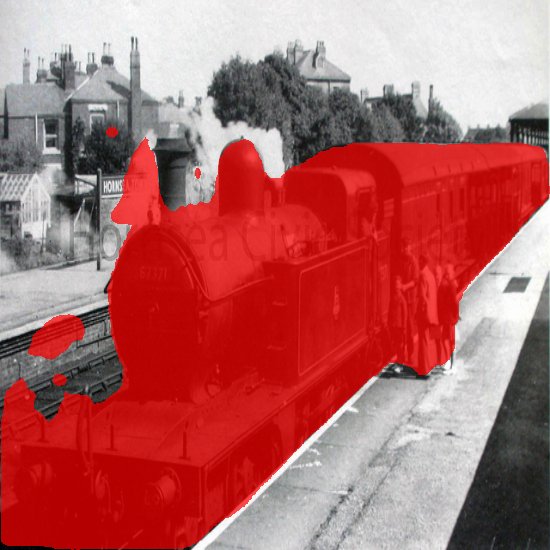}\\
    \includegraphics[width=0.33\linewidth]{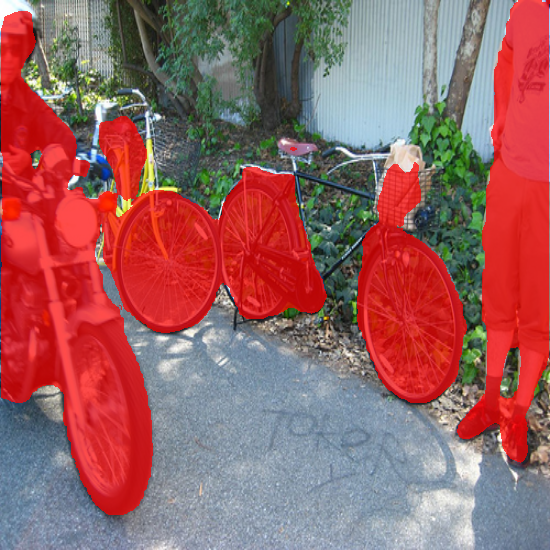}\includegraphics[width=0.33\linewidth]{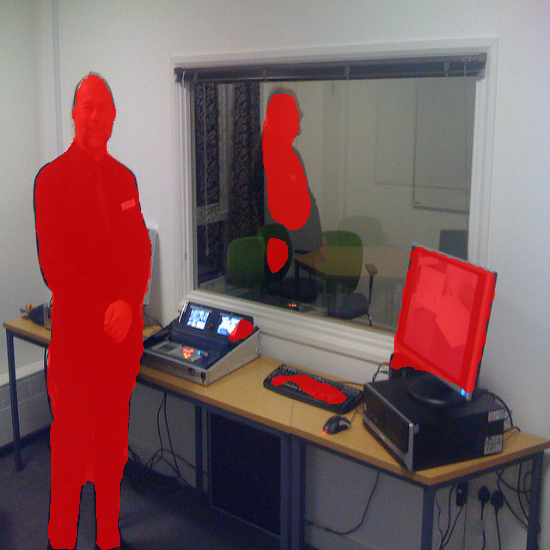}\includegraphics[width=0.33\linewidth]{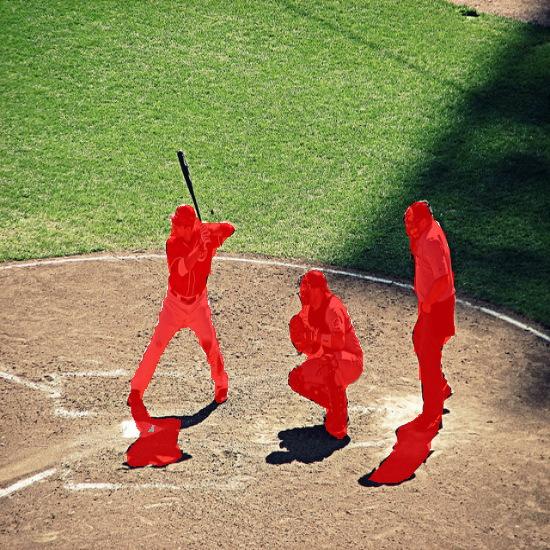}\\
    \includegraphics[width=0.33\linewidth]{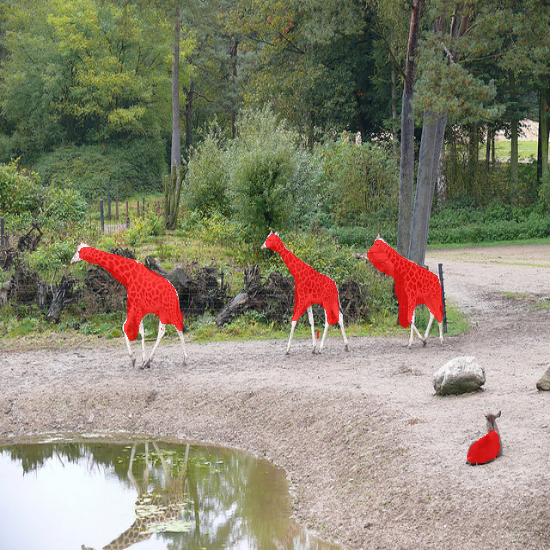}\includegraphics[width=0.33\linewidth]{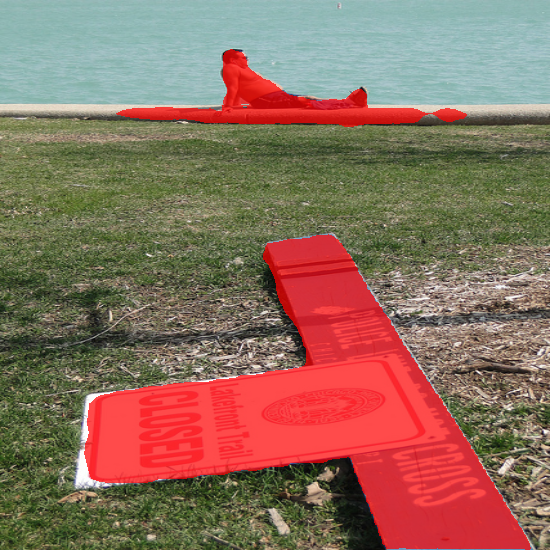}\includegraphics[width=0.33\linewidth]{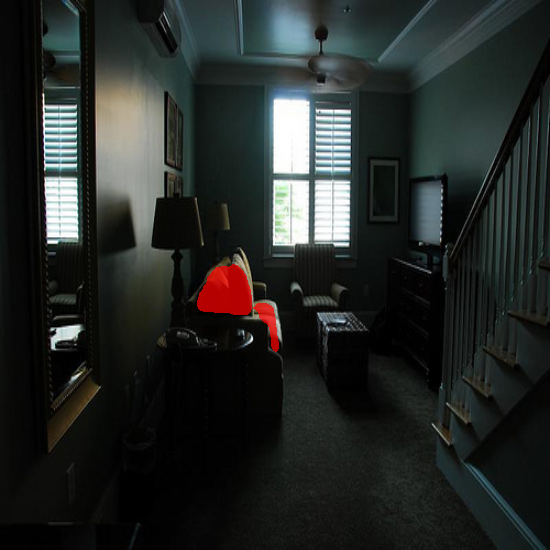}
\end{center}
   \caption{Our method predicts accurate distance maps that are used to separate the foreground from the background (first row). However, it has difficulty with very thin parts that belong to objects, such as the bicycle frame or the legs of the giraffes (first column). Our method also has difficulty with object reflections and object shadows (second row), as well as with poorly lit scenes (third row).}
\label{fig:gb}
\end{figure*}

Furthermore, in Fig.~\ref{fig:gb} we show that our method predicts accurate distance maps. These are used to approximate a signed distance function where the positive set of function values describe the foreground segmentation. We also show cases that our method has difficulty with, such as object reflections, object shadows, low-light scenes, and very thin object parts.

\begin{table}[t]
    \caption{Comparing foreground segmentation between YOLACT~\cite{Bolya_2019_ICCV} and our approach on COCO \texttt{val2017}~\cite{lin2014microsoft}. Both with a ResNet50~\cite{He2016DeepRL} backbone.}
    \label{tab:iou}
    \begin{center}
        \begin{tabular}{ cccc } 
            \hline
            Method & $z$ & topk & mIoU  \\ 
            \hline
            YOLACT~\cite{Bolya_2019_ICCV} & - & 5 & 0.66 \\ 
             & - & 50 & 0.65 \\ 
             & - & 100 & 0.64 \\ 
            \hline
            \textbf{Ours} & \textbf{0.0} & \textbf{-} & \textbf{0.70} \\ 
             & 3.0 & - & 0.67 \\ 
             & 5.0 & - & 0.63 \\ 
             & 10.0 & - & 0.53 \\ 
            \hline
        \end{tabular}
    \end{center}
\end{table}

\section{CONCLUSION AND FUTURE WORK}
We explored a new direction for the instance segmentation problem, where the predictions made by a neural network has a direct geometric interpretation. That is, we train the network to predict distances to object contours that are used to compute a signed distance function. We analyse this approach and show that it outperforms the state-of-the-art instance segmentation YOLACT at foreground-background segmentation. A possible explanation is that some shapes are not well described by anchor-boxes.

Although the predicted distance maps contain information on the individual instances it remains challenging to map this representation to a full instance segmentation. For this representation, an object can consist of different segmentation blobs that are separated due to e.g. occlusion. To label these object related parts with the same label identification is one of the challenges with instance segmentation, even though the blobs fit well to the object scope. We therefore believe this representation to be a promising direction for instance segmentation methods.


